\documentclass[conference]{IEEEtran}
\IEEEoverridecommandlockouts
\usepackage{cite}
\usepackage{amsmath,amssymb,amsfonts}
\usepackage{algorithmic}
\usepackage{graphicx}
\usepackage{textcomp}
\usepackage{subcaption}
\usepackage[dvipsnames]{xcolor}
\usepackage{multirow}
\usepackage{stfloats}
\usepackage{booktabs}
\usepackage{listings}
\usepackage[most]{tcolorbox}
\usepackage{pifont}
\usepackage{pgfplots}
\pgfplotsset{compat=1.18}
\usetikzlibrary{decorations.pathreplacing, shadows}

\lstdefinestyle{pyz3}{
    language=Python,
    basicstyle=\ttfamily\footnotesize,
    keywordstyle=\color{codepurple},
    stringstyle=\color{codegreen},
    commentstyle=\color{codegreen},
    showstringspaces=false,
    breaklines=false,
    frame=none,
    aboveskip=0pt,
    belowskip=0pt,
    xleftmargin=0pt,
    columns=fullflexible,
    keepspaces=true,
    escapeinside={(*@}{@*)},
    morekeywords={def,return},
}

\def\BibTeX{{\rm B\kern-.05em{\sc i\kern-.025em b}\kern-.08em
    T\kern-.1667em\lower.7ex\hbox{E}\kern-.125emX}}

\definecolor{codegreen}{HTML}{008000}
\definecolor{codeblue}{HTML}{0000FF}
\definecolor{codeyellow}{HTML}{795E26}
\definecolor{codepurple}{HTML}{AF00DB}
\definecolor{codered}{HTML}{CD3131}
\definecolor{codecyan}{HTML}{267F99}

\newcommand{\ours}{BODHI}

\begin{document}

\title{
\ours: Precise OS Kernel Specification Inference
}

\author{\IEEEauthorblockN{Zhiming Chang}
\IEEEauthorblockA{\textit{Department of Applied Mathematics and Statistics} \\
\textit{Johns Hopkins University}\\
Baltimore, US \\
zchang9@jh.edu}
\and
\IEEEauthorblockN{Ziyang Li}
\IEEEauthorblockA{\textit{Department of Computer Science} \\
\textit{Johns Hopkins University}\\
Baltimore, US \\
ziyang@cs.jhu.edu}
}

\maketitle

\begin{abstract}
The formal verification of operating system kernels requires precise specifications that capture the intended behavior of system calls. 
Writing these specifications manually demands deep domain expertise, motivating the use of large language models (LLMs) to automate the process. 
However, in OSV-Bench, a benchmark of 245 specification generation tasks derived from the Hyperkernel OS kernel, the best reported Pass@1 is 55.10\% \cite{b1}. 
We propose a domain knowledge prompting method (\ours), which augments the standard few-shot prompt with a structured C-to-Python translation guide covering 15 categories of domain-specific translation patterns. 
Inspired by Structured Chain-of-Thought (SCoT) prompting \cite{b2}, the guide organizes translation by separation of concerns, addressing pre-condition extraction and post-condition generation as distinct categories. 
Evaluated on nine models from six providers (Anthropic, Mistral, Amazon, DeepSeek, Meta, Alibaba), covering dense, mixture-of-experts and reasoning architectures, \ours\ improves every model tested, with gains ranging from $+$11\% to $+$32\%. 
The best configuration (Claude Opus 4.6 + \ours) reaches 96.73\% Pass@1. 
\ours\ reduces both syntax and semantic errors, with the strongest effect on models that have sufficient instruction-following capability to utilize structured reference material. 
These results demonstrate that domain knowledge injection is a model-agnostic technique that substantially bridges the gap between general-purpose code generation and formal specification synthesis.
\end{abstract}

\section{Introduction}

Formal software verification uses rigorous mathematical reasoning to prove that a system is free of bugs, which is crucial for safety-critical systems, such as operating system kernels, where software errors can have disastrous consequences \cite{b3}. 
However, constructing formal specifications, a prerequisite for verification, remains a bottleneck that requires deep domain expertise. 
For example, updating the seL4 microkernel to support efficient RPC modified less than 5\% of the code, yet the resulting specification and proof adjustments took approximately one person-year of effort \cite{b4}.

Large language models have shown strong capabilities in code generation \cite{b5, b6}, but their performance in generating formal specifications remains limited. 
OSV-Bench \cite{b1}, currently the largest benchmark for OS kernel specification generation, evaluated 12 large language models on 245 tasks derived from the Hyperkernel operating system \cite{b7}. 
Each task requires converting a C system call implementation into a Python/Z3 state-machine specification. 
The best performing model achieved only 55.10\% Pass@1, exposing a significant gap between natural language code generation and formal specification synthesis.

This gap results from several challenges inherent in the specification generation process: 

\begin{itemize}

    \item \textbf{Long-context complexity:} Each task prompt contains approximately 26K tokens, including a programming model, a small number of examples, a functional description and code for the target system call.
    
    \item \textbf{Semantic gap:} C kernel code must be converted to the declarative Z3 \cite{b8} Python specification, and these two programming styles are fundamentally different.
    
    \item \textbf{Interleaved concerns:} In C code, pre-conditions (error-checking logic) and post-conditions (state-changing logic) are structurally mixed, but they must be clearly distinguished in the specification.
    
    \item \textbf{Domain-specific knowledge:} Kernel API patterns such as page table entry (PTE) formulas, reference count semantics, TLB flush operations and IPC state-machines are highly specialized and not covered by general-purpose LLM training.
    
\end{itemize}

We adopt a simple but effective approach to address these challenges: we provide a comprehensive \textit{structured C-to-Python translation guide}, building upon a standard set of prompts with a small number of examples. 
We call this approach \ours. 
The guide is a 519-line reference document divided into 15 categories, offering clear translation patterns for every major construct encountered in the Hyperkernel domain. 
Drawing on the principles of Structured Chain-of-Thought (SCoT) prompting \cite{b2}, it organizes the translation task into distinct concern categories (pre-condition extraction, post-condition patterns, field access syntax, and domain-specific formulas) rather than requiring the model to infer all patterns from a few examples alone.

In this paper, we make the following contributions:

\begin{itemize}

    \item A domain-specific translation guide for OSV-Bench specification generation, covering 15 categories of C-to-Python/Z3 translation patterns derived from analysis of verified ground truth specifications.
    
    \item Comprehensive evaluation of \ours\ on nine models from six providers with isolated API calls per-task, demonstrating model-agnostic effectiveness across diverse architectures.
    
    \item Detailed analysis showing that \ours\ reduces both syntax and semantic errors, with the largest gains on mid-tier models that have sufficient instruction-following ability but incomplete domain knowledge.
    
    \item New state-of-the-art on the OSV-Bench at 96.73\% Pass@1, substantially exceeding the previous best result of 55.10\%.
    
\end{itemize}

\section{Background}

\subsection{OSV-Bench}\label{sec:prompt}

The OSV-Bench \cite{b1} is a benchmark for evaluating LLMs in specification generation for OS kernel verification. 
It is built on Hyperkernel \cite{b7}, a formally verified operating system kernel that uses the Z3 SMT solver \cite{b8} for verification.

Each of the 245 tasks requires generating a Python/Z3 state-machine specification for a system call. 
The specification defines pre-conditions and post-conditions. 
The task prompt includes: 
(1) a system verification assumption document, 
(2) a programming model defining available classes, constants, and functions, 
(3) few-shot examples of system call specifications, 
and (4) the target system call's functional description and (potentially buggy) C implementation.

The 245 tasks cover 49 system calls with varying bug injections. 
Each task's C implementation may contain one or more of five buggy types: incorrect pointer operations, incorrect privilege checks, memory leaks, buffer overflows, and missing bounds checks. 
In addition, 49 tasks contain correct implementations. 
A specification is correct if and only if the Z3 verifier produces results consistent with the oracle specification across all kernel implementation variants.

Specifications are verified in a Docker environment containing the Hyperkernel build system and Z3 solver \cite{b1}, following the verification pipeline described in Section~\ref{sec:z3}.
The primary metric is Pass@1: the fraction of tasks in which the LLM-generated specification passes verification on the first attempt.

The published OSV-Bench evaluation covers 12 LLMs, with the best Pass@1 reaching 55.10\% (Doubao-1.5-pro) \cite{b1}.
Notably, reasoning-enhanced models (o1, DeepSeek-R1) do not outperform standard instruction-following models, suggesting that specification generation requires domain-specific knowledge rather than general reasoning capability.


\subsection{Z3 and the Verification Pipeline}\label{sec:z3}

The specifications in OSV-Bench are written using the Python API of Z3 \cite{b8}, a Satisfiability Modulo Theories (SMT) solver.
SMT solvers determine the satisfiability of logical formulas over theories such as fixed-width bitvectors and arrays, which directly model hardware registers, memory pages and kernel data structures in operating system kernels.

Each specification is a Python function that encodes the intended behavior of one system call as a state-machine transition.
Formally, a specification for system call $s$ is a function
\begin{equation}\label{eq:spec}
f_s(\mathbf{S}, \mathbf{a}) \rightarrow (\phi, \mathbf{S}')
\end{equation}
where $\mathbf{S}$ denotes the current kernel state (a collection of maps and scalar fields), $\mathbf{a}$ denotes the system call arguments, $\phi$ is the \textit{pre-condition} (a Z3 Boolean expression representing the conjunction of all validity checks), and $\mathbf{S}'$ is the \textit{post-state}.
The post-state satisfies $\mathbf{S}' = \texttt{update}(\mathbf{S})$ when $\phi$ holds, and $\mathbf{S}' = \mathbf{S}$ otherwise.

\begin{figure}[t]
    \centering
    \begin{tikzpicture}[
            cbox/.style={draw=black!40, rounded corners=2pt, minimum height=18pt,
                inner sep=3pt, font=\sffamily\small,
                fill={rgb,255:red,251;green,251;blue,251},
                drop shadow={shadow xshift=0.4pt, shadow yshift=-0.4pt, fill=black!15}},
            implbox/.style={cbox, draw={rgb,255:red,128;green,0;blue,0},
                fill={rgb,255:red,128;green,0;blue,0}, text=white, font=\sffamily\small\bfseries},
            specbox/.style={cbox, draw={rgb,255:red,0;green,51;blue,102},
                fill={rgb,255:red,0;green,51;blue,102}, text=white, font=\sffamily\small\bfseries},
            ltoolbox/.style={cbox, draw=red!30!black!40, fill=red!6, text=red!60!black!40, font=\sffamily\scriptsize},
            solverbox/.style={cbox, draw=orange, fill=orange, text=white, font=\sffamily\small\bfseries},
            resbox/.style={cbox, draw=green!50!black, fill=green!5, font=\sffamily\footnotesize},
            eqlbl/.style={font=\sffamily\scriptsize, orange!80!black},
            arr_l/.style={->, >=stealth, line width=0.9pt, red!40!black!50},
            arr_r/.style={->, >=stealth, line width=0.9pt, blue!40!black!40},
            arr_c/.style={->, >=stealth, line width=0.9pt, orange!60},
            lbl_l/.style={font=\sffamily\scriptsize, red!40!black!50},
            lbl_r/.style={font=\sffamily\scriptsize, blue!40!black!40},
        ]

        \node[implbox, text width=2.8cm, align=center] (c) at (-2.2, 0) {C Implementation};
        \node[ltoolbox, text width=2.0cm, align=center] (ir) at (-2.2, -1.1) {LLVM IR};
        \node[ltoolbox, text width=2.0cm, align=center] (behav) at (-2.2, -2.4) {Implementation\\[-1pt]Behavior};

        \draw[arr_l] (c) -- node[right, lbl_l] {compile} (ir);
        \draw[arr_l] (ir) -- node[right, lbl_l] {symbolic exec.} (behav);

        \node[specbox, text width=4.6cm, align=center] (spec) at (2.2, 0) {Specification (Python/Z3)};

        \node[solverbox, minimum width=8.0cm] (z3) at (0.4, -3.6) {Z3 Solver};
        \draw[arr_l] (behav) -- node[right, lbl_l] {formula} (z3.north -| behav);
        \draw[arr_r] (spec.south) -- ++(0, -2.35) node[right, lbl_r, midway] {constraints} -- (z3.north -| spec);

        \node[resbox, draw=green!50!black, fill=green!5, text width=3.0cm, align=center] (verified) at (-2.0, -4.7)
            {\textcolor{green!50!black}{\bfseries\checkmark~Verified}};
        \node[cbox, draw=red!60!black, fill=red!5, text width=3.0cm, align=center, font=\sffamily\footnotesize] (counter) at (2.8, -4.7)
            {\textcolor{red!70!black}{\bfseries\texttimes~Counterexample}};

        \draw[arr_c] (z3.south -| verified) -- node[right, eqlbl] {equivalent} (verified);
        \draw[arr_c] (z3.south -| counter) -- node[right, eqlbl] {divergent} (counter);

    \end{tikzpicture}
    \caption{Hyperkernel verification pipeline. The C implementation (left) is compiled to LLVM IR and symbolically executed to extract a behavioral formula. The specification (right) provides Z3 constraints. Z3 checks consistency: if the formula satisfies the constraints for all inputs, the system call is verified; otherwise, a counterexample is produced.
    }
    \label{fig:verify_pipeline}
\end{figure}
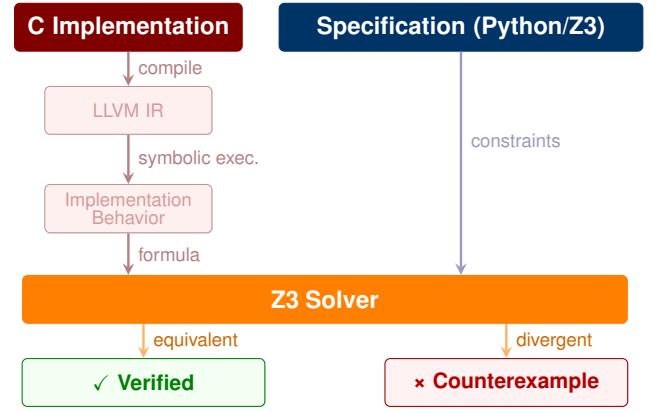

Fig.~\ref{fig:verify_pipeline} illustrates the verification pipeline used by Hyperkernel.
The C implementation is compiled to LLVM intermediate representation (IR) \cite{b9}, and the toolchain performs symbolic execution on the IR, extracting a logical formula that captures the implementation's behavior over all possible inputs.
Denoting the implementation's extracted behavior as $g_s(\mathbf{S}, \mathbf{a})$, Z3 checks behavioral equivalence:
\begin{equation}\label{eq:verify}
\text{verify}(s) \iff \forall\, \mathbf{a}:\; g_s(\mathbf{S}, \mathbf{a}) \equiv f_s(\mathbf{S}, \mathbf{a})
\end{equation}
If a discrepancy exists, Z3 produces a \textit{counterexample}: a concrete $\mathbf{a}^*$ such that $g_s(\mathbf{S}, \mathbf{a}^*) \not\equiv f_s(\mathbf{S}, \mathbf{a}^*)$.
If no such $\mathbf{a}^*$ exists, the system call is verified correct.

This \textit{push-button} approach requires no manual proofs: given a correct specification, verification is fully automated.
The bottleneck therefore shifts from proof construction to specification writing, precisely the task that OSV-Bench evaluates and that \ours\ addresses.

\subsection{Structured Chain-of-Thought (SCoT)}

Chain-of-Thought (CoT) prompting \cite{b10} improves LLM reasoning by eliciting intermediate reasoning steps before the final answer.
Structured Chain-of-Thought (SCoT) \cite{b2} generalizes the Chain-of-Thought framework to code generation, where sequential, branching, and looping program constructs are utilized to direct the intermediate reasoning procedure.
Rather than generating free-form reasoning text, SCoT first produces a pseudocode skeleton that captures the control flow of the target program, then fills in the implementation details within that skeleton.
This structural decomposition of the \textit{output generation process} has shown improvements over basic CoT on code generation benchmarks such as HumanEval and MBPP \cite{b2}.
We draw inspiration from SCoT but adapt it to the specification generation setting in a fundamentally different way.
SCoT decomposes the \textit{output} by structuring how the model generates code, while \ours\ decomposes the \textit{input} by organizing domain-specific translation patterns into structured reference material (Section~\ref{sec:method}).

\newsavebox{\specbox}
\begin{lrbox}{\specbox}
\begin{minipage}{6.2cm}
\begin{lstlisting}[style=pyz3]
(*@\color{codepurple}def@*) (*@\color{codeyellow}sys\_set\_runnable@*)(old, pid):
    cond = (*@\color{codeyellow}z3.And@*)(
        (*@\color{codeyellow}z3.And@*)(pid > 0, pid < (*@\color{codecyan}dt@*).NPROC),
        old.procs[pid].ppid == old.current,
        old.procs[pid].state ==
            (*@\color{codecyan}dt@*).proc_state.PROC_EMBRYO,
    )

    new = old.(*@\color{codeyellow}copy@*)()
    (*@\color{codeblue}new@*).procs[pid].state =
        (*@\color{codecyan}dt@*).proc_state.PROC_RUNNABLE

    (*@\color{codepurple}return@*) cond, (*@\color{codeyellow}util.If@*)(cond, new, old)
\end{lstlisting}
\end{minipage}
\end{lrbox}

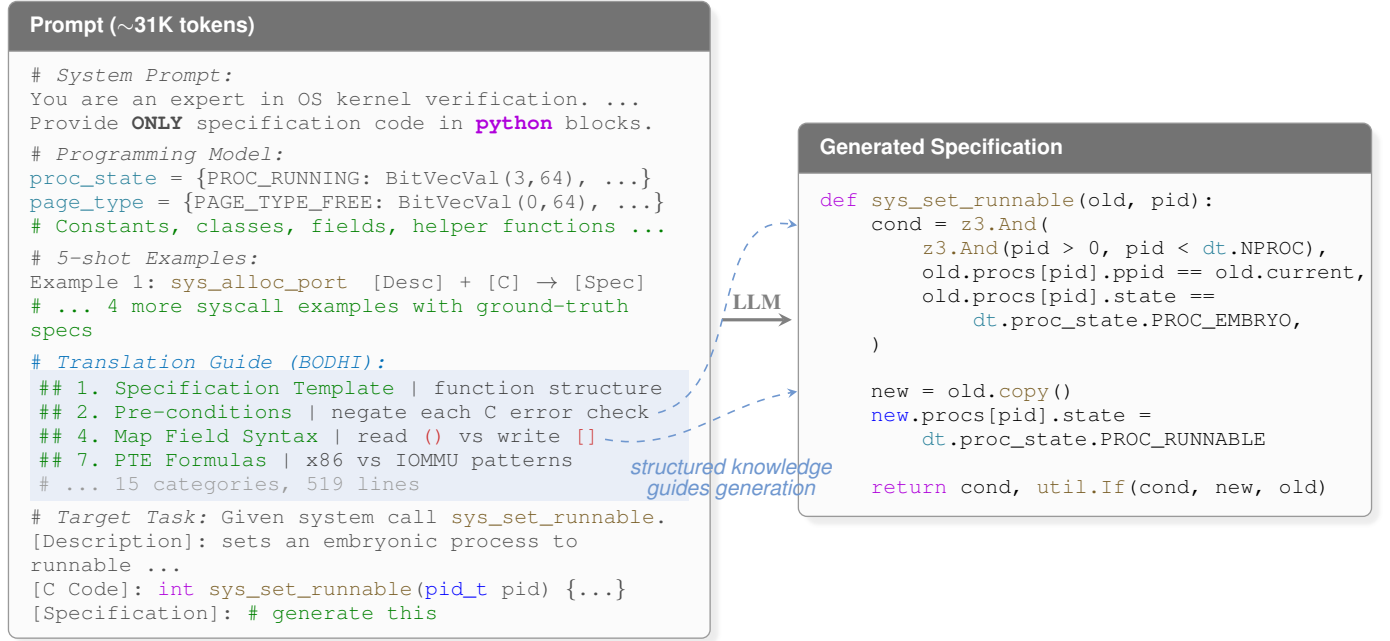
\begin{figure*}[htbp]
    \centering
    \begin{tikzpicture}[
            arrow/.style={->, >=stealth, line width=1.2pt},
            dasharrow/.style={->, >=stealth, dashed, line width=0.6pt},
            leftbrace/.style={decorate, decoration={brace, amplitude=3pt, mirror, raise=2pt}},
            rightbrace/.style={decorate, decoration={brace, amplitude=3pt, raise=2pt}},
        ]

        \node[inner sep=0pt] (input) at (-4.5, 0) {%
        \begin{tcolorbox}[
                enhanced, boxrule=0.5pt, arc=3pt,
                colframe=black!40, colbacktitle=black!55,
                colback={rgb,255:red,251;green,251;blue,251},
                coltitle=white, fonttitle=\sffamily\bfseries\footnotesize,
                toptitle=2.5pt, bottomtitle=2.5pt,
                left=5pt, right=5pt, top=3pt, bottom=3pt,
                drop fuzzy shadow=black!15!white,
                title={Prompt ($\sim$31K tokens)},
                width=9.3cm
            ]
            \ttfamily\footnotesize\color{black!70}%
            \#\ \textit{System Prompt:}\\
            You are an expert in OS kernel verification. ... Provide \textbf{ONLY} specification code in \textbf{\textcolor{codepurple}{python}} blocks.\\[3pt]
            \#\ \textit{Programming Model:}\\
            \textcolor{codecyan}{proc\_state} = \{PROC\_RUNNING: BitVecVal(3,64), ...\}\\
            \textcolor{codecyan}{page\_type} = \{PAGE\_TYPE\_FREE: BitVecVal(0,64), ...\}\\
            \textcolor{codegreen}{\# Constants, classes, fields, helper functions ...}\\[3pt]
            \#\ \textit{5-shot Examples:}\\
            Example 1: \textcolor{codeyellow}{sys\_alloc\_port}\quad [Desc] + [C] $\to$ [Spec]\\
            \textcolor{codegreen}{\# ... 4 more syscall examples with ground-truth specs}\\[3pt]
            \textcolor{NavyBlue}{\#\ \textit{Translation Guide (\ours):}}
            \colorbox{NavyBlue!8}{\parbox{\dimexpr\linewidth-6pt}{%
            \color{black!75}%
            \textcolor{codegreen}{\#\# 1.\ Specification Template} — function structure\\
            \textcolor{codegreen}{\#\# 2.\ Pre-conditions} — negate each C error check\\
            \textcolor{codegreen}{\#\# 4.\ Map Field Syntax} — read \textcolor{codered}{()} vs write \textcolor{codered}{[]}\\
            \textcolor{codegreen}{\#\# 7.\ PTE Formulas} — x86 vs IOMMU patterns\\
            \textcolor{black!40}{\# ... 15 categories, 519 lines}%
            }}\\[3pt]
            \#\ \textit{Target Task:}
            Given system call \textcolor{codeyellow}{sys\_set\_runnable}.\\
            {[}Description{]}: sets an embryonic process to runnable ...\\
            {[}C Code{]}: \textcolor{codepurple}{int} \textcolor{codeyellow}{sys\_set\_runnable}(\textcolor{codeblue}{pid\_t} pid) \{...\}\\
            {[}Specification{]}: \textcolor{codegreen}{\# generate this}%
        \end{tcolorbox}%
        };


        \node[inner sep=0pt] (output) at (5.1, 0) {%
        \begin{tcolorbox}[
                enhanced, boxrule=0.5pt, arc=3pt,
                colframe=black!40, colbacktitle=black!55,
                colback={rgb,255:red,251;green,251;blue,251},
                coltitle=white, fonttitle=\sffamily\bfseries\footnotesize,
                toptitle=2.5pt, bottomtitle=2.5pt,
                left=5pt, right=5pt, top=3pt, bottom=3pt,
                drop fuzzy shadow=black!15!white,
                title={Generated Specification},
                width=7.6cm
            ]
            \usebox{\specbox}
        \end{tcolorbox}%
        };

        \draw[arrow, black!50] ([xshift=4pt]input.east) -- node[above, font=\footnotesize\bfseries] {LLM} ([xshift=-2pt]output.west);

        \draw[dasharrow, NavyBlue!60] ([xshift=-20pt, yshift=-35pt]input.east) to[out=15, in=170] ([yshift=110pt]output.south west);
        \draw[dasharrow, NavyBlue!60] ([xshift=-40pt, yshift=-45pt]input.east) to[out=-10, in=190] ([yshift=-27pt]output.west);

        \node[font=\sffamily\footnotesize\itshape, NavyBlue!60, align=center] at ([xshift=-25pt, yshift=-60pt]output.west) {structured knowledge\\[-1pt]guides generation};

    \end{tikzpicture}
    \caption{The \ours\ prompting pipeline. The prompt (left) contains five components; the \textit{translation guide} (highlighted, our contribution) provides 15 categories of structured domain knowledge between the few-shot examples and the target task. Dashed arrows show how structured guide categories directly shape the generated specification (right): Category~2 guides pre-condition translation (red), Category~4 guides field access syntax (blue).}
    \label{fig:prompt}
\end{figure*}

\section{Methodology}\label{sec:method}

Our method, \ours, augments the standard few-shot prompt with a structured \textit{C-to-Python Translation Guide}. 
This section describes the prompt structure, the guide's design and construction, and the key principle of separation of concerns.

\subsection{Prompt Structure}

Fig.~\ref{fig:prompt} illustrates the prompt composition for the baseline and \ours\ configurations.
Both configurations share four prompt components inherited from the OSV-Bench evaluation protocol \cite{b1}:
\textit{(1)} a \textit{system prompt} (approximately 2K tokens) that instructs the LLM to generate Python/Z3 specifications;
\textit{(2)} the \textit{programming model} (approximately 10K tokens), which defines the kernel state $\mathbf{S}$ (Eq.~\ref{eq:spec}) including map declarations, bitvector widths and accessor conventions;
\textit{(3)} five expert-selected \textit{few-shot examples} (approximately 12K tokens), each pairing a C system call with its oracle Z3 specification; 
and \textit{(4)} the \textit{target task} (approximately 2K tokens), comprising the natural-language functional description and the C implementation to be translated.
\ours\ adds a fifth component: the \textit{translation guide} (approximately 5K tokens), inserted between the few-shot examples and the target task.

The placement of the guide is deliberate.
By positioning it after the few-shot examples and immediately before the target task, the guide occupies the most recent context when the model begins generation.
This leverages the recency effect in transformer attention \cite{b17}: tokens closer to the generation boundary receive higher attention weight, ensuring that domain-specific patterns are maximally available during specification synthesis.
The few-shot examples, placed earlier, establish the general input-output format, while the guide refines the model's knowledge of domain-specific translation rules just before the model applies them.

The prompt is submitted as a single API call with no multi-turn interaction.
The LLM generates the complete specification in one pass and the output is directly submitted to the Z3 verifier.
This design keeps the method simple and reproducible: the only variable between the baseline and \ours\ is the presence of the translation guide.

\subsection{Translation Guide Design}

The translation guide is a 519-line reference document covering 15 categories of domain-specific translation patterns (TABLE~\ref{tab1}). 

\begin{table*}[htbp]
    \caption{Translation guide categories. Each category targets a specific class of specification errors.}
    \begin{center}
        \begin{tabular}{cllll}
            \toprule
            \textbf{ID} & \textbf{Category} & \textbf{Content} & \textbf{Error class} & \textbf{Addresses}\\
            \midrule
            1 & Specification template & Standard function structure, special cases & Structural & Missing return format\\
            2 & Pre-condition translation & C error checks $\rightarrow$ Z3 negated conditions & Semantic & Incorrect guard logic\\
            3 & Post-condition patterns & \texttt{util.If} usage, \texttt{new.*} conventions & Syntax & Wrong conditional API\\
            4 & Map field syntax & Read \texttt{()} vs. write \texttt{[]} dual syntax & Syntax & Most common syntax error\\
            5 & Operator rules & Z3 Python API operator overloading & Syntax & Z3 API misuse\\
            6 & Constant prefixes & All constants use \texttt{dt.*} namespace & Syntax & Namespace violations\\
            7 & Page table PTE formulas & x86 vs. IOMMU, 7 mapping variants & Semantic & Core domain knowledge\\
            8 & Shadow metadata & 5 shadow fields for page table mappings & Semantic & Implicit state omission\\
            9 & Reference counting & Ownership transfer refcnt updates & Semantic & Hidden in C, explicit in spec\\
            10 & TLB flush & x86 vs. IOMMU flush call differences & Completeness & Omitted operations\\
            11 & State pointers & 6 \texttt{\_ptr\_to\_int} field exact names & Syntax & Naming errors\\
            12 & Field name mapping & C variable $\rightarrow$ Z3 path correspondence & Semantic & Translation accuracy\\
            13 & C helper functions & Helpers requiring inline expansion & Semantic & Abstraction mismatch\\
            14 & Available helpers & 9 predefined helper functions & Syntax & API boundary\\
            15 & IPC system calls & send/call/reply\_wait/extintr patterns & Completeness & Most complex syscall type\\
            \bottomrule
        \end{tabular}
        \label{tab1}
    \end{center}
\end{table*}

It was constructed through an iterative process: 
\textit{(1)} run baseline prompting on a subset of tasks, 
\textit{(2)} classify failures into error types, 
\textit{(3)} identify recurring patterns in the ground truth oracle specifications that address each error type, 
and \textit{(4)} codify these patterns into structured guide categories. 
Each category targets a specific class of specification errors observed in preliminary experiments.

We used Claude Haiku 4.5 as the development model to maximize failure coverage during guide construction.
The Claude Haiku 4.5 baseline achieved 50.20\% Pass@1 (123/245), producing 49 syntax errors and 73 semantic errors across 122 failed tasks spanning 30 of the 49 system calls.
This broad failure surface, nearly double the 16 failed system calls observed with Claude Opus 4.6, ensured that the guide would address error patterns common to models of varying capability levels.
By designing the guide against a mid-tier model's failures, the resulting categories are inclusive: patterns that a weaker model needs are a superset of those that a stronger model lacks, enabling the guide to generalize upward to more capable models as confirmed by the evaluation in Section~\ref{sec:results}.
Manual inspection of the 122 failures revealed strong clustering around three subsystems (IPC, IOMMU and resource reclamation), motivating the three-tier organization of the guide.

The 15 categories were organized into three tiers based on the error analysis.
The first tier targets \textit{syntax-level} patterns that ensure $\phi$ and $\mathbf{S}'$ (Eq.~\ref{eq:spec}) can be correctly constructed: specification template, pre-condition translation, post-condition patterns, map field syntax, operator rules and constant prefixes.
These address the 49 syntax errors observed in the Haiku baseline, such as using parentheses instead of brackets for map field writes and omitting the \texttt{dt.} namespace prefix for constants.
The second tier targets \textit{domain-specific} semantic patterns: PTE formulas, shadow metadata, reference counting, TLB flush, state pointer names and field name mapping.
These address the semantic errors concentrated in IOMMU and resource reclamation system calls, where the model consistently lacked specialized translation knowledge.
The third tier targets \textit{completeness} in the most complex system calls: C helper function expansion, available helper function inventory and IPC state machine patterns.
These address the IPC and interrupt remapping failures, where even correct syntax and domain knowledge are insufficient without explicit multi-step coordination patterns.
The resulting guide comprises 519 lines.

The 15 categories cover all recurring error patterns identified in the development set; the remaining failures are unique to individual system calls with no extractable general pattern.
Importantly, the guide contains only general translation rules and illustrative patterns derived from the oracle specifications. It includes no task-specific code or complete solutions for any system call, functioning as a domain textbook rather than an answer key.


Fig.~\ref{fig:guide_example} shows an excerpt from two representative guide categories. 
\begin{figure}[htbp]
    \centering
    
    \begin{tcolorbox}[
            enhanced,
            boxrule=0.5pt, arc=3pt,
            colframe={rgb,255:red,128;green,0;blue,0},
            colbacktitle={rgb,255:red,128;green,0;blue,0},
            colback={rgb,255:red,251;green,251;blue,251},
            coltitle=white,
            fonttitle=\sffamily\bfseries\footnotesize,
            toptitle=2.5pt, bottomtitle=2.5pt,
            left=4pt, right=4pt, top=3pt, bottom=4pt,
            drop fuzzy shadow=black!15!white,
            title={Category 2: Pre-condition Translation}
        ]
        {\ttfamily\footnotesize\color{codegreen}\itshape // Each C error return $\to$ negation in cond}\\[3pt]
        \renewcommand{\arraystretch}{1.35}
        \fontsize{7.5pt}{9pt}\selectfont
        \begin{tabular}{@{}p{0.45\linewidth}@{\hspace{4pt}}c@{\hspace{4pt}}p{0.44\linewidth}@{}}
            \sffamily\textit{\color{black!60}C source} & & \sffamily\textit{\color{black!60}Z3 guard} \\
            \midrule
            \texttt{\textcolor{codepurple}{if} (\textcolor{codered}{!}\textcolor{codeyellow}{is\_pid\_valid}(pid))}\newline
            \texttt{\ \ \textcolor{codepurple}{\ \ return} \textcolor{codered}{-E};}
            & \textcolor{codegreen}{$\rightarrow$}
            & \texttt{\textcolor{codeyellow}{is\_pid\_valid}(pid)} \\[2pt]
            \texttt{\textcolor{codepurple}{if} (state \textcolor{codered}{!=} PROC\_RUNNING)}\newline
            \texttt{\ \ \textcolor{codepurple}{\ \ return} \textcolor{codered}{-E};}
            & \textcolor{codegreen}{$\rightarrow$}
            & \texttt{old.procs[pid].state}\newline\texttt{\ \textcolor{codeblue}{==} \textcolor{codecyan}{dt.proc\_state}.}\newline\texttt{\ PROC\_RUNNING} \\[2pt]
            \texttt{\textcolor{codepurple}{if} (n \textcolor{codered}{>=} NPAGE)}\newline
            \texttt{\ \ \textcolor{codepurple}{\ \ return} \textcolor{codered}{-E};}
            & \textcolor{codegreen}{$\rightarrow$}
            & \texttt{\textcolor{codeyellow}{z3.ULT}(n, \textcolor{codecyan}{dt}.NPAGE)} \\
            \bottomrule
        \end{tabular}
    \end{tcolorbox}
    
    \vspace{4pt}
    
    \begin{tcolorbox}[
            enhanced,
            boxrule=0.5pt, arc=3pt,
            colframe={rgb,255:red,0;green,51;blue,102},
            colbacktitle={rgb,255:red,0;green,51;blue,102},
            colback={rgb,255:red,251;green,251;blue,251},
            coltitle=white,
            fonttitle=\sffamily\bfseries\footnotesize,
            toptitle=2.5pt, bottomtitle=2.5pt,
            left=4pt, right=4pt, top=3pt, bottom=4pt,
            drop fuzzy shadow=black!15!white,
            title={Category 4: Map Field Syntax}
        ]
        {\ttfamily\footnotesize\color{codegreen}\itshape // Read with () \quad vs.\quad Write with []}\\[3pt]
        \renewcommand{\arraystretch}{1.35}
        \fontsize{7.5pt}{9pt}\selectfont
        \begin{tabular}{@{}p{0.47\linewidth}@{\hspace{4pt}}p{0.48\linewidth}@{}}
            \sffamily\textit{\color{black!60}Read access} & \sffamily\textit{\color{black!60}Write access} \\
            \midrule
            \texttt{old.pages[pn].data\textcolor{codered}{\textbf{(}}idx\textcolor{codered}{\textbf{)}}}
            & \texttt{\textcolor{codeblue}{new}.pages[pn].data\textcolor{codered}{\textbf{[}}idx\textcolor{codered}{\textbf{]}} = val} \\[2pt]
            \texttt{old.procs[pid].ofile\textcolor{codered}{\textbf{(}}fd\textcolor{codered}{\textbf{)}}}
            & \texttt{\textcolor{codeblue}{new}.procs[pid].ofile\textcolor{codered}{\textbf{[}}fd\textcolor{codered}{\textbf{]}} = val} \\[2pt]
            \texttt{old.procs[pid].nr\_pages\textcolor{codered}{\textbf{()}}}
            & \texttt{\textcolor{codeblue}{new}.procs[pid].nr\_pages\textcolor{codered}{\textbf{[}}pn\textcolor{codered}{\textbf{]}} += 1} \\
            \bottomrule
        \end{tabular}
    \end{tcolorbox}
    
    \caption{Excerpts from two translation guide categories. \textit{Category~2}: each C error check is negated to form the Z3 pre-condition guard. \textit{Category~4}: Z3 map fields use parentheses for reads but brackets for writes, the most common syntax error in baseline experiments.}
    \label{fig:guide_example}
\end{figure}

The pre-condition category teaches the LLM that each C error check must be negated when translated to a Z3 guard condition. 
The map field syntax category addresses the most common syntax error observed in baseline experiments: Z3's dual-access convention, where reading uses parentheses, but writing uses brackets.

\subsection{Separation of Concerns}

A key design principle of the guide is the \textit{separation of concerns}: patterns for the pre-condition $\phi$ (error check negation) and the post-state $\mathbf{S}'$ (state updates) are organized into distinct categories (Eq.~\ref{eq:spec}).
This reflects a structural property of kernel system calls, where error-checking logic and state-mutation logic are naturally decoupled in the C source code.

Fig.~\ref{fig:c2py} illustrates this with a real Hyperkernel system call (\texttt{sys\_set\_runnable}). 
The C implementation interleaves pre-conditions and post-conditions in a single control flow, but the Z3 specification separates them into two distinct regions: a \texttt{z3.And()} block for $\phi$ and a sequential assignment block for $\mathbf{S}'$.
The guide teaches the LLM to perform this structural transformation by providing separate pattern references for each region.

\newsavebox{\ccodebox}
\begin{lrbox}{\ccodebox}
\begin{minipage}{6.4cm}
\begin{lstlisting}[style=pyz3, language=C, morekeywords={int,struct,if,return}]
(*@\color{codeblue}int@*) (*@\color{codeyellow}sys\_set\_runnable@*)((*@\color{codeblue}pid\_t@*) pid) {
    (*@\color{codeblue}struct@*) proc *proc;

    (*@\color{codepurple}if@*) ((*@\color{codered}!@*)(*@\color{codeyellow}is\_pid\_valid@*)(pid))
        (*@\color{codepurple}return@*) (*@\color{codered}-ESRCH@*);
    proc = (*@\color{codeyellow}get\_proc@*)(pid);
    (*@\color{codepurple}if@*) (proc(*@\color{codered}->@*)ppid (*@\color{codered}!=@*) current)
        (*@\color{codepurple}return@*) (*@\color{codered}-EACCES@*);
    (*@\color{codepurple}if@*) (proc(*@\color{codered}->@*)state (*@\color{codered}!=@*) (*@\color{codecyan}PROC\_EMBRYO@*))
        (*@\color{codepurple}return@*) (*@\color{codered}-EINVAL@*);

    (*@\color{codeblue}proc->state@*) = (*@\color{codecyan}PROC\_RUNNABLE@*);
    (*@\color{codeyellow}proc\_ready\_add@*)(proc);
    (*@\color{codepurple}return@*) 0;
}
\end{lstlisting}
\end{minipage}
\end{lrbox}

\newsavebox{\pycodebox}
\begin{lrbox}{\pycodebox}
\begin{minipage}{6.4cm}
\begin{lstlisting}[style=pyz3]
(*@\color{codepurple}def@*) (*@\color{codeyellow}sys\_set\_runnable@*)(old, pid):
    cond = (*@\color{codeyellow}z3.And@*)(
        (*@\color{codeyellow}z3.And@*)(pid (*@\color{codered}>@*) 0, pid (*@\color{codered}<@*) (*@\color{codecyan}dt@*).NPROC),
        old.procs[pid].ppid (*@\color{codeblue}==@*) old.current,
        old.procs[pid].state (*@\color{codeblue}==@*)
            (*@\color{codecyan}dt@*).proc_state.PROC_EMBRYO)

    new = old.(*@\color{codeyellow}copy@*)()
    (*@\color{codeblue}new@*).procs[pid].state =
        (*@\color{codecyan}dt@*).proc_state.PROC_RUNNABLE
    (*@\color{codepurple}return@*) cond, (*@\color{codeyellow}util.If@*)(cond, new, old)
\end{lstlisting}
\end{minipage}
\end{lrbox}

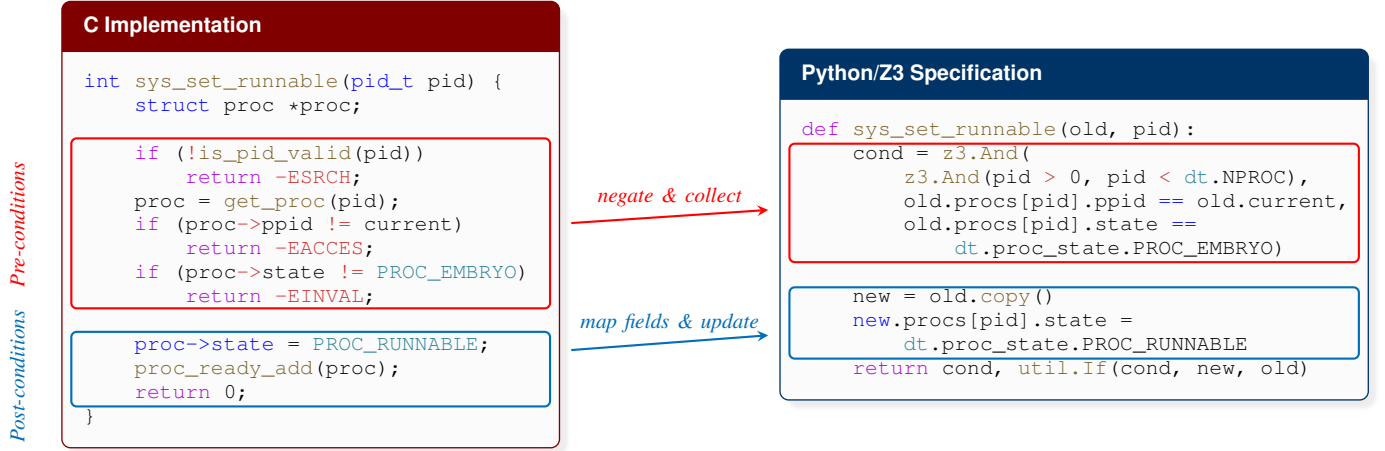
\begin{figure*}[htbp]
    \centering
    \begin{tikzpicture}[
            highlight/.style={draw, rounded corners=2pt, inner sep=3pt, line width=0.8pt},
            arrow/.style={->, >=stealth, thick, shorten >=4pt, shorten <=4pt},
        ]

        \node[inner sep=0pt] (cbox) at (-5.6, 3.0) {%
        \begin{tcolorbox}[
                enhanced, boxrule=0.5pt, arc=3pt,
                colframe={rgb,255:red,128;green,0;blue,0},
                colbacktitle={rgb,255:red,128;green,0;blue,0},
                colback={rgb,255:red,251;green,251;blue,251},
                coltitle=white, fonttitle=\sffamily\bfseries\footnotesize,
                toptitle=2.5pt, bottomtitle=2.5pt,
                left=5pt, right=5pt, top=4pt, bottom=4pt,
                drop fuzzy shadow=black!15!white,
                title={C Implementation},
                width=6.6cm
            ]
            \usebox{\ccodebox}
        \end{tcolorbox}%
        };

        \draw[highlight, red, fill opacity=0.5]
            ([xshift=4pt, yshift=-52pt]cbox.north west) rectangle ([xshift=-4pt, yshift=14pt]cbox.east |- 0,1.4);
        \draw[highlight, NavyBlue, fill opacity=0.5]
            ([xshift=4pt, yshift=8pt]cbox.west |- 0,1.3) rectangle ([xshift=-4pt, yshift=16pt]cbox.south east);
        \node[font=\footnotesize\itshape, red, rotate=90, anchor=south] at ([xshift=-10pt]cbox.west |- 0,3.0) {Pre-conditions};
        \node[font=\footnotesize\itshape, NavyBlue, rotate=90, anchor=south] at ([xshift=-10pt]cbox.west |- 0,1.0) {Post-conditions};

        \node[inner sep=0pt] (pybox) at (4.5, 3.0) {%
        \begin{tcolorbox}[
                enhanced, boxrule=0.5pt, arc=3pt,
                colframe={rgb,255:red,0;green,51;blue,102},
                colbacktitle={rgb,255:red,0;green,51;blue,102},
                colback={rgb,255:red,251;green,251;blue,251},
                coltitle=white, fonttitle=\sffamily\bfseries\footnotesize,
                toptitle=2.5pt, bottomtitle=2.5pt,
                left=5pt, right=5pt, top=4pt, bottom=4pt,
                drop fuzzy shadow=black!15!white,
                title={Python/Z3 Specification},
                width=7.8cm
            ]
            \usebox{\pycodebox}
        \end{tcolorbox}%
        };

        \draw[highlight, red, fill opacity=0.5]
            ([xshift=4pt, yshift=-36pt]pybox.north west) rectangle ([xshift=-4pt, yshift=20pt]pybox.east |- 0,1.8);
        \draw[highlight, NavyBlue, fill opacity=0.5]
            ([xshift=4pt, yshift=2pt]pybox.west |- 0,2.1) rectangle ([xshift=-4pt, yshift=16pt]pybox.south east);

        \draw[arrow, red] ([yshift=6pt]cbox.east |- 0,2.8) -- node[above, font=\footnotesize\itshape, red] {negate \& collect} ([yshift=20pt]pybox.west |- 0,2.5);
        \draw[arrow, NavyBlue] ([yshift=-2pt]cbox.east |- 0,1.4) -- node[above, font=\footnotesize\itshape, NavyBlue] {map fields \& update} ([yshift=7pt]pybox.west |- 0,1.3);
    \end{tikzpicture}
    \caption{Separation of concerns in the C-to-Z3 translation (\texttt{sys\_set\_runnable}). Red overlay: pre-conditions (C error checks are negated and collected into a \texttt{z3.And()} guard). Blue overlay: post-conditions (C state mutations are mapped to Z3 field assignments).
    }
    \label{fig:c2py}
\end{figure*}

This separation is not merely organizational. 
In baseline experiments without the guide, the most common semantic errors involved the LLM incorrectly mixing pre-condition and post-condition logic. 
For example, applying negation to a state update expression or omitting an error check because it appeared syntactically similar to a state mutation. 
By structuring the guide around this separation, \ours\ provides the LLM with distinct mental models for each translation task, reducing such cross-contamination errors.

\section{Evaluation}

We design our evaluation to answer four research questions: 
Does the translation guide improve specification generation accuracy across diverse models and providers? 
What types of error does the guide reduce? 
How does the guide's contribution compare to model capability improvements? 
Which system call categories benefit the most from the guide?

\subsection{Evaluation Setup}

\textbf{Baseline:} 
Our baseline follows the OSV-Bench evaluation protocol \cite{b1}: a system prompt instructing the LLM to generate Python/Z3 specifications(2K tokens), followed by the definition of the programming model (10K tokens), five expert-selected few-shot examples (12K tokens), and the functional description of the target task and the implementation of C (2K tokens). 
The LLM generates the complete specification in a single call.

\textbf{Models:} 
We evaluate nine models from six providers: Claude Opus 4.6, Claude Sonnet 4, and Claude Haiku 4.5 (Anthropic); Devstral 2 123B (Mistral); Nova Premier (Amazon); DeepSeek-V3.2 and DeepSeek-R1 (DeepSeek); Llama 4 Maverick 17B-A3B (Meta); and Qwen3 80B-A3B (Alibaba). 
This selection spans dense architectures, mixture-of-expert models and a reasoning-optimized model. 
The evaluations use greedy decoding (temperature 0.0) and \texttt{max\_tokens} 4096, following the OSV-Bench prompt structure and verification pipeline \cite{b1}.
Unlike the original evaluation, which averages three runs, we report single-run results; with greedy decoding, repeated runs are deterministic for most providers.
DeepSeek-R1, as a reasoning model, does not support temperature configuration.

\textbf{Task isolation:} 
Each of the 245 tasks is processed as an independent API call with a fresh request body. 
No conversation history, system state or prior outputs are shared between tasks. 
This isolation is critical because multiple tasks share the same system call (with different bug injections) and leakage could inflate performance.

\textbf{Token consumption:} 
Each task prompt comprises approximately 26K input tokens for the baseline and approximately 31K for the \ours\ version. 
The output specifications average approximately 340 tokens. 
The total token consumption in all 18 experiment configurations (9 models $\times$ 2 methods $\times$ 245 tasks) is approximately 126M tokens.

\textbf{Verification:} 
The generated specifications are verified using the OSV-Bench Docker environment, which contains the Hyperkernel build system and Z3 solver (v4.5.0) \cite{b1}. 
The verifier performs symbolic execution on compiled LLVM IR and checks behavioral equivalence (Eq.~\ref{eq:verify}) with oracle specifications across all kernel implementation variants.

\textbf{Evaluation Metrics:}
We adopt Pass@1 as the primary metric: each task is submitted once and the specification either passes or fails Z3 verification (Eq.~\ref{eq:verify}). 
We also classify failures into \textit{syntax errors} (the specification raises a Python or Z3 type error before verification completes) and \textit{semantic errors} (the specification executes but produces verification results inconsistent with the oracle). 
This three-way split (pass / syntax / semantic) enables fine-grained analysis of how \ours\ affects different error modes.

\subsection{Main Results}\label{sec:results}

\begin{figure*}[htbp]
    \centering
    \footnotesize
    \begin{tikzpicture}
        \begin{axis}[
                ybar,
                width=\textwidth,
                height=6.0cm,
                bar width=14pt,
                ylabel={Pass@1 (\%)},
                symbolic x coords={DS-R1, Qwen3, Llama4, DS-V3, Nova, Devstral2, Haiku, Sonnet, Opus},
                xtick=data,
                x tick label style={font=\footnotesize, rotate=0},
                ymin=0, ymax=120,
                ytick={0,20,40,60,80,100},
                nodes near coords,
                every node near coord/.append style={font=\footnotesize\bfseries, rotate=90, anchor=west, xshift=0pt, yshift=-2pt},
                legend style={at={(0,1)}, anchor=north west, font=\footnotesize, legend columns=2, column sep=5pt},
                grid=major,
                grid style={dashed, gray!30},
                enlarge x limits=0.08,
            ]
            \addplot[fill=black!15, draw=black!40] coordinates {
                (DS-R1, 26.53) (Qwen3, 38.78) (Llama4, 38.78) (DS-V3, 38.78)
                (Nova, 43.67) (Devstral2, 48.16) (Haiku, 50.20) (Sonnet, 61.22) (Opus, 74.69)};
            \addplot[fill=NavyBlue!50, draw=NavyBlue] coordinates {
                (DS-R1, 40.41) (Qwen3, 49.80) (Llama4, 55.10) (DS-V3, 56.73)
                (Nova, 58.78) (Devstral2, 80.00) (Haiku, 76.33) (Sonnet, 75.10) (Opus, 96.73)};
            \legend{Baseline, + \ours\ (Ours)}
            
            \draw[red!80!black, dashed, thick] ({rel axis cs:0,0} |- {axis cs:DS-R1,55.10}) -- ({rel axis cs:1,0} |- {axis cs:DS-R1,55.10});
            \node[red!80!black, font=\footnotesize, anchor=south west] at ({rel axis cs:0.01,0} |- {axis cs:DS-R1,60.10}) {prev.\ best 55.10\%};
            
        \end{axis}
    \end{tikzpicture}
    \caption{Pass@1 on OSV-Bench across nine models from six providers, sorted by baseline performance. Each pair compares baseline (gray) with \ours\ (blue). Dashed red line: previous state-of-the-art (Doubao-1.5-pro \cite{b1}). \ours\ improves every model with significant gains.}
    \label{fig:pass1}
\end{figure*}
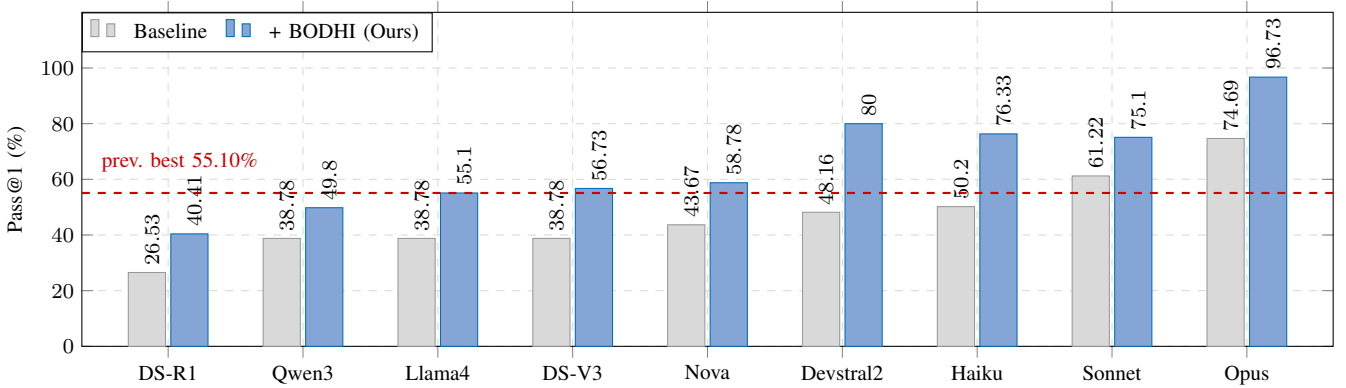

Fig.~\ref{fig:pass1} and TABLE~\ref{tab2} present our main results in nine models along with the published OSV-Bench baselines.

\begin{table*}[htbp]
    \caption{Pass@1 (\%) on OSV-Bench. Published baselines from \cite{b1}; our results use isolated per-task API calls with temperature 0.0. All nine models show consistent improvement with \ours.}
    \begin{center}
    \setlength{\tabcolsep}{5.0pt}
    \begin{tabular}{l ccccccc}
    \toprule
    \textbf{Model} & Incorr. Ptr. & Incorr. Priv. & Mem. Leak & Buf. Ovfl. & Bounds Chk. & Correct & \textbf{Total}\\
    \midrule
    \multicolumn{8}{l}{\textit{Published baselines} \cite{b1}}\\
    Doubao-1.5-pro & 50.70 & 48.21 & 45.95 & 40.74 & 52.78 & 63.27 & 55.10\\
    DeepSeek-Chat & 38.02 & 39.29 & 36.49 & 44.44 & 43.52 & 51.02 & 46.53\\
    Claude Sonnet 3.5 & 39.44 & 41.96 & 39.19 & 48.15 & 39.81 & 46.94 & 44.90\\
    \midrule
    \multicolumn{8}{l}{\textit{Our baselines (without \ours)}}\\
    Claude Opus 4.6 & 70.42 & 72.32 & 60.81 & 62.96 & 69.44 & 79.59 & 74.69\\
    Claude Sonnet 4 & 54.93 & 51.79 & 52.70 & 46.30 & 55.56 & 75.51 & 61.22\\
    Claude Haiku 4.5 & 39.44 & 41.96 & 41.89 & 42.59 & 43.52 & 59.18 & 50.20\\
    Devstral 2 & 46.48 & 41.96 & 47.30 & 42.59 & 48.15 & 51.02 & 48.16\\
    Nova Premier & 39.44 & 35.71 & 37.84 & 27.78 & 40.74 & 48.98 & 43.67\\
    DeepSeek-V3.2 & 30.99 & 36.61 & 31.08 & 29.63 & 34.26 & 38.78 & 38.78\\
    Llama 4 Maverick & 33.80 & 33.04 & 25.68 & 35.19 & 40.74 & 42.86 & 38.78\\
    Qwen3 & 43.66 & 29.46 & 32.43 & 20.37 & 33.33 & 44.90 & 38.78\\
    DeepSeek-R1$^\dagger$ & 21.13 & 22.32 & 18.92 & 29.63 & 28.70 & 24.49 & 26.53\\
    \midrule
    \multicolumn{8}{l}{\textit{Our results (with \ours)}}\\
    Claude Opus 4.6 & \textbf{97.18} {\color{ForestGreen}\scriptsize($+$26.8)} & \textbf{95.54} {\color{ForestGreen}\scriptsize($+$23.2)} & \textbf{94.59} {\color{ForestGreen}\scriptsize($+$33.8)} & \textbf{98.15} {\color{ForestGreen}\scriptsize($+$35.2)} & \textbf{96.30} {\color{ForestGreen}\scriptsize($+$26.9)} & \textbf{97.96} {\color{ForestGreen}\scriptsize($+$18.4)} & \textbf{96.73} {\color{ForestGreen}\scriptsize($+$22.0)}\\
    Claude Sonnet 4 & 77.46 {\color{ForestGreen}\scriptsize($+$22.5)} & 66.96 {\color{ForestGreen}\scriptsize($+$15.2)} & 75.68 {\color{ForestGreen}\scriptsize($+$23.0)} & 79.63 {\color{ForestGreen}\scriptsize($+$33.3)} & 70.37 {\color{ForestGreen}\scriptsize($+$14.8)} & 85.71 {\color{ForestGreen}\scriptsize($+$10.2)} & 75.10 {\color{ForestGreen}\scriptsize($+$13.9)}\\
    Claude Haiku 4.5 & 76.06 {\color{ForestGreen}\scriptsize($+$36.6)} & 66.07 {\color{ForestGreen}\scriptsize($+$24.1)} & 70.27 {\color{ForestGreen}\scriptsize($+$28.4)} & 75.93 {\color{ForestGreen}\scriptsize($+$33.3)} & 72.22 {\color{ForestGreen}\scriptsize($+$28.7)} & 83.67 {\color{ForestGreen}\scriptsize($+$24.5)} & 76.33 {\color{ForestGreen}\scriptsize($+$26.1)}\\
    Devstral 2 & 74.65 {\color{ForestGreen}\scriptsize($+$28.2)} & 72.32 {\color{ForestGreen}\scriptsize($+$30.4)} & 75.68 {\color{ForestGreen}\scriptsize($+$28.4)} & 85.19 {\color{ForestGreen}\scriptsize($+$42.6)} & 77.78 {\color{ForestGreen}\scriptsize($+$29.6)} & 85.71 {\color{ForestGreen}\scriptsize($+$34.7)} & 80.00 \textbf{\color{ForestGreen}\scriptsize($+$31.8)}\\
    Nova Premier & 56.34 {\color{ForestGreen}\scriptsize($+$16.9)} & 53.57 {\color{ForestGreen}\scriptsize($+$17.9)} & 52.70 {\color{ForestGreen}\scriptsize($+$14.9)} & 37.04 {\color{ForestGreen}\scriptsize($+$9.3)} & 57.41 {\color{ForestGreen}\scriptsize($+$16.7)} & 63.27 {\color{ForestGreen}\scriptsize($+$14.3)} & 58.78 {\color{ForestGreen}\scriptsize($+$15.1)}\\
    DeepSeek-V3.2 & 47.89 {\color{ForestGreen}\scriptsize($+$16.9)} & 53.57 {\color{ForestGreen}\scriptsize($+$17.0)} & 51.35 {\color{ForestGreen}\scriptsize($+$20.3)} & 59.26 {\color{ForestGreen}\scriptsize($+$29.6)} & 52.78 {\color{ForestGreen}\scriptsize($+$18.5)} & 63.27 {\color{ForestGreen}\scriptsize($+$24.5)} & 56.73 {\color{ForestGreen}\scriptsize($+$18.0)}\\
    Llama 4 Maverick & 52.11 {\color{ForestGreen}\scriptsize($+$18.3)} & 48.21 {\color{ForestGreen}\scriptsize($+$15.2)} & 39.19 {\color{ForestGreen}\scriptsize($+$13.5)} & 46.30 {\color{ForestGreen}\scriptsize($+$11.1)} & 57.41 {\color{ForestGreen}\scriptsize($+$16.7)} & 59.18 {\color{ForestGreen}\scriptsize($+$16.3)} & 55.10 {\color{ForestGreen}\scriptsize($+$16.3)}\\
    Qwen3 & 54.93 {\color{ForestGreen}\scriptsize($+$11.3)} & 37.50 {\color{ForestGreen}\scriptsize($+$8.0)} & 44.59 {\color{ForestGreen}\scriptsize($+$12.2)} & 38.89 {\color{ForestGreen}\scriptsize($+$18.5)} & 47.22 {\color{ForestGreen}\scriptsize($+$13.9)} & 55.10 {\color{ForestGreen}\scriptsize($+$10.2)} & 49.80 {\color{ForestGreen}\scriptsize($+$11.0)}\\
    DeepSeek-R1$^\dagger$ & 35.21 {\color{ForestGreen}\scriptsize($+$14.1)} & 33.93 {\color{ForestGreen}\scriptsize($+$11.6)} & 27.03 {\color{ForestGreen}\scriptsize($+$8.1)} & 44.44 {\color{ForestGreen}\scriptsize($+$14.8)} & 37.04 {\color{ForestGreen}\scriptsize($+$8.3)} & 53.06 {\color{ForestGreen}\scriptsize($+$28.6)} & 40.41 {\color{ForestGreen}\scriptsize($+$13.9)}\\
    \bottomrule
    \multicolumn{8}{l}{\parbox{0.95\textwidth}{\footnotesize $^\dagger$Reasoning model; temperature not configurable. High syntax error rate (59\%/53\%) due to output format non-compliance.\\
Our baseline (26.53\%) is lower than the published result (40.00\% \cite{b1}), likely due to API version differences and single-run evaluation.}}
    \end{tabular}
    \label{tab2}
    \end{center}
\end{table*}

\ours\ improves Pass@1 on every model tested, with gains ranging from $+$11.0 (Qwen3) to $+$31.8 (Devstral 2) percentage points, demonstrating that structured domain knowledge injection is effective regardless of the model provider, architecture or capability level.

Improvements are not monotonically related to baseline capability.
Devstral 2 achieves the largest gain ($+$31.8\%), reaching 80.0\% with \ours\ and surpassing all baselines except Claude Opus 4.6. 
Claude Haiku 4.5 and Claude Opus 4.6 follow. 
The three weakest baselines (DeepSeek-R1, Qwen3, Llama 4 Maverick) show moderate but consistent improvements of $+$11-16\%. 
This suggests that the effectiveness of \ours\ depends on a model's ability to utilize structured reference material: models with sufficient instruction-following capability benefit substantially, while models limited by output format compliance (DeepSeek-R1) or fundamental code generation capability derive smaller but still positive gains.

Six models with \ours\ surpass the previous state-of-the-art (Doubao-1.5-pro, 55.10\% \cite{b1}): Claude Opus 4.6, Devstral 2, Claude Haiku 4.5, Claude Sonnet 4, Nova Premier and DeepSeek-V3.2; Llama 4 Maverick matches it. 
The best configuration reaches 96.73\%, failing on only 8 of 245 tasks. 
The guide improves performance across all bug categories for all models, with the greatest gains appearing in the buffer overflow and memory leak categories, where domain-specific PTE formulas and reference counting patterns are the most relevant.

\subsection{Error Analysis}

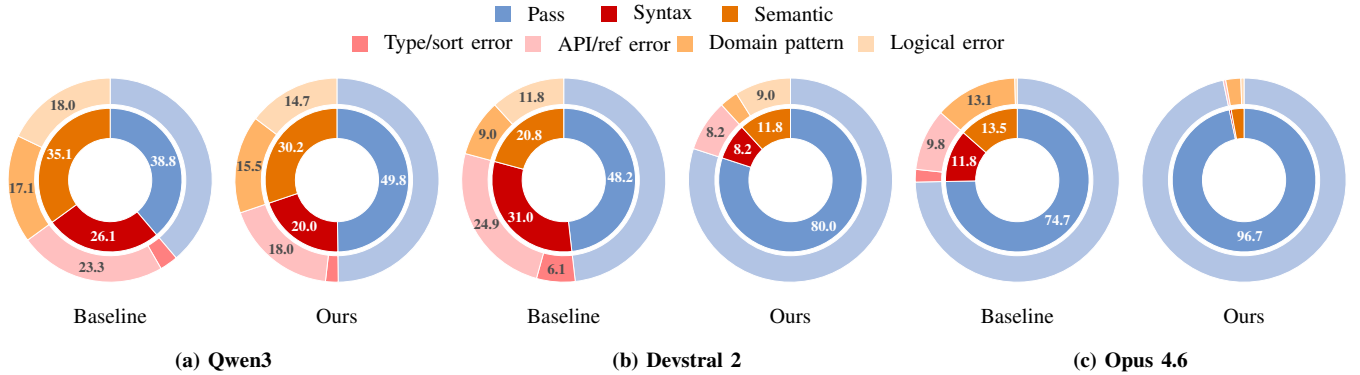
\begin{figure*}[htbp]
    \centering
    \begin{tikzpicture}

    \begin{scope}[shift={(7.5,2.1)}]
        \fill[NavyBlue!60] (-2.4,0) rectangle (-2.2,0.2); \node[font=\footnotesize, anchor=west] at (-2.1,0.1) {Pass};
        \fill[red!80!black] (-1.0,0) rectangle (-0.8,0.2); \node[font=\footnotesize, anchor=west] at (-0.7,0.1) {Syntax};
        \fill[orange!90!black] (0.6,0) rectangle (0.8,0.2); \node[font=\footnotesize, anchor=west] at (0.9,0.1) {Semantic};
        \fill[red!50] (-4.3,-0.4) rectangle (-4.1,-0.2); \node[font=\footnotesize, anchor=west] at (-4.0,-0.3) {Type/sort error};
        \fill[red!25] (-2.0,-0.4) rectangle (-1.8,-0.2); \node[font=\footnotesize, anchor=west] at (-1.7,-0.3) {API/ref error};
        \fill[orange!60] (0.0,-0.4) rectangle (0.2,-0.2); \node[font=\footnotesize, anchor=west] at (0.3,-0.3) {Domain pattern};
        \fill[orange!30] (2.4,-0.4) rectangle (2.6,-0.2); \node[font=\footnotesize, anchor=west] at (2.7,-0.3) {Logical error};
    \end{scope}

    \def\ringseg#1#2#3#4#5{\fill[#5, draw=white, line width=0.3pt] (#1:#3) arc (#1:#2:#3) -- (#2:#4) arc (#2:#1:#4) -- cycle;}
    \newcommand{\ringlabel}[6]{%
        \pgfmathsetmacro{\midang}{(#1+#2)/2}%
        \pgfmathsetmacro{\midr}{(#3+#4)/2}%
        \node[font=\fontsize{5.5pt}{6pt}\selectfont\bfseries, #6] at (\midang:\midr) {#5};%
    }

    \begin{scope}[shift={(0,0)}]
        \ringseg{90}{-49.7}{0.55}{0.95}{NavyBlue!60}
        \ringseg{-49.7}{-144.0}{0.55}{0.95}{red!80!black}
        \ringseg{-144.0}{-270}{0.55}{0.95}{orange!90!black}
        \ringseg{90}{-49.7}{1.0}{1.35}{NavyBlue!30}
        \ringseg{-49.7}{-60.1}{1.0}{1.35}{red!50}
        \ringseg{-60.1}{-144.0}{1.0}{1.35}{red!25}
        \ringseg{-144.0}{-205.6}{1.0}{1.35}{orange!60}
        \ringseg{-205.6}{-270}{1.0}{1.35}{orange!30}
        \ringlabel{90}{-49.7}{0.55}{0.95}{38.8}{white}
        \ringlabel{-49.7}{-144.0}{0.55}{0.95}{26.1}{white}
        \ringlabel{-144.0}{-270}{0.55}{0.95}{35.1}{white}
        \ringlabel{-60.1}{-144.0}{1.0}{1.35}{23.3}{black!70}
        \ringlabel{-144.0}{-205.6}{1.0}{1.35}{17.1}{black!70}
        \ringlabel{-205.6}{-270}{1.0}{1.35}{18.0}{black!70}
        \node[font=\footnotesize, black] at (0,-1.8) {Baseline};
    \end{scope}

    \begin{scope}[shift={(3.0,0)}]
        \ringseg{90}{-89.3}{0.55}{0.95}{NavyBlue!60}
        \ringseg{-89.3}{-161.3}{0.55}{0.95}{red!80!black}
        \ringseg{-161.3}{-270}{0.55}{0.95}{orange!90!black}
        \ringseg{90}{-89.3}{1.0}{1.35}{NavyBlue!30}
        \ringseg{-89.3}{-96.5}{1.0}{1.35}{red!50}
        \ringseg{-96.5}{-161.3}{1.0}{1.35}{red!25}
        \ringseg{-161.3}{-217.1}{1.0}{1.35}{orange!60}
        \ringseg{-217.1}{-270}{1.0}{1.35}{orange!30}
        \ringlabel{90}{-89.3}{0.55}{0.95}{49.8}{white}
        \ringlabel{-89.3}{-161.3}{0.55}{0.95}{20.0}{white}
        \ringlabel{-161.3}{-270}{0.55}{0.95}{30.2}{white}
        \ringlabel{-96.5}{-161.3}{1.0}{1.35}{18.0}{black!70}
        \ringlabel{-161.3}{-217.1}{1.0}{1.35}{15.5}{black!70}
        \ringlabel{-217.1}{-270}{1.0}{1.35}{14.7}{black!70}
        \node[font=\footnotesize, black] at (0,-1.8) {Ours};
    \end{scope}

    \node[font=\footnotesize\bfseries] at (1.5,-2.4) {(a) Qwen3};

    \begin{scope}[shift={(6.0,0)}]
        \ringseg{90}{-83.5}{0.55}{0.95}{NavyBlue!60}
        \ringseg{-83.5}{-195.1}{0.55}{0.95}{red!80!black}
        \ringseg{-195.1}{-270}{0.55}{0.95}{orange!90!black}
        \ringseg{90}{-83.5}{1.0}{1.35}{NavyBlue!30}
        \ringseg{-83.5}{-105.5}{1.0}{1.35}{red!50}
        \ringseg{-105.5}{-195.1}{1.0}{1.35}{red!25}
        \ringseg{-195.1}{-227.5}{1.0}{1.35}{orange!60}
        \ringseg{-227.5}{-270}{1.0}{1.35}{orange!30}
        \ringlabel{90}{-83.5}{0.55}{0.95}{48.2}{white}
        \ringlabel{-83.5}{-195.1}{0.55}{0.95}{31.0}{white}
        \ringlabel{-195.1}{-270}{0.55}{0.95}{20.8}{white}
        \ringlabel{-83.5}{-105.5}{1.0}{1.35}{6.1}{black!70}
        \ringlabel{-105.5}{-195.1}{1.0}{1.35}{24.9}{black!70}
        \ringlabel{-195.1}{-227.5}{1.0}{1.35}{9.0}{black!70}
        \ringlabel{-227.5}{-270}{1.0}{1.35}{11.8}{black!70}
        \node[font=\footnotesize, black] at (0,-1.8) {Baseline};
    \end{scope}

    \begin{scope}[shift={(9.0,0)}]
        \ringseg{90}{-198.0}{0.55}{0.95}{NavyBlue!60}
        \ringseg{-198.0}{-227.5}{0.55}{0.95}{red!80!black}
        \ringseg{-227.5}{-270}{0.55}{0.95}{orange!90!black}
        \ringseg{90}{-198.0}{1.0}{1.35}{NavyBlue!30}
        \ringseg{-198.0}{-227.5}{1.0}{1.35}{red!25}
        \ringseg{-227.5}{-238.0}{1.0}{1.35}{orange!60}
        \ringseg{-238.0}{-270}{1.0}{1.35}{orange!30}
        \ringlabel{90}{-198.0}{0.55}{0.95}{80.0}{white}
        \ringlabel{-198.0}{-227.5}{0.55}{0.95}{8.2}{white}
        \ringlabel{-227.5}{-270}{0.55}{0.95}{11.8}{white}
        \ringlabel{-198.0}{-227.5}{1.0}{1.35}{8.2}{black!70}
        \ringlabel{-238.0}{-270}{1.0}{1.35}{9.0}{black!70}
        \node[font=\footnotesize, black] at (0,-1.8) {Ours};
    \end{scope}

    \node[font=\footnotesize\bfseries] at (7.5,-2.4) {(b) Devstral 2};

    \begin{scope}[shift={(12.0,0)}]
        \ringseg{90}{-178.9}{0.55}{0.95}{NavyBlue!60}
        \ringseg{-178.9}{-221.4}{0.55}{0.95}{red!80!black}
        \ringseg{-221.4}{-270}{0.55}{0.95}{orange!90!black}
        \ringseg{90}{-178.9}{1.0}{1.35}{NavyBlue!30}
        \ringseg{-178.9}{-186.1}{1.0}{1.35}{red!50}
        \ringseg{-186.1}{-221.4}{1.0}{1.35}{red!25}
        \ringseg{-221.4}{-268.6}{1.0}{1.35}{orange!60}
        \ringseg{-268.6}{-270}{1.0}{1.35}{orange!30}
        \ringlabel{90}{-178.9}{0.55}{0.95}{74.7}{white}
        \ringlabel{-178.9}{-221.4}{0.55}{0.95}{11.8}{white}
        \ringlabel{-221.4}{-270}{0.55}{0.95}{13.5}{white}
        \ringlabel{-186.1}{-221.4}{1.0}{1.35}{9.8}{black!70}
        \ringlabel{-221.4}{-268.6}{1.0}{1.35}{13.1}{black!70}
        \node[font=\footnotesize, black] at (0,-1.8) {Baseline};
    \end{scope}

    \begin{scope}[shift={(15.0,0)}]
        \ringseg{90}{-258.1}{0.55}{0.95}{NavyBlue!60}
        \ringseg{-258.1}{-259.6}{0.55}{0.95}{red!80!black}
        \ringseg{-259.6}{-270}{0.55}{0.95}{orange!90!black}
        \ringseg{90}{-258.1}{1.0}{1.35}{NavyBlue!30}
        \ringseg{-258.1}{-259.6}{1.0}{1.35}{red!25}
        \ringseg{-259.6}{-268.2}{1.0}{1.35}{orange!60}
        \ringseg{-268.2}{-270}{1.0}{1.35}{orange!30}
        \ringlabel{90}{-258.1}{0.55}{0.95}{96.7}{white}
        \node[font=\footnotesize, black] at (0,-1.8) {Ours};
    \end{scope}

    \node[font=\footnotesize\bfseries] at (13.5,-2.4) {(c) Opus 4.6};

    \end{tikzpicture}

    \caption{Error distribution as nested donut charts for three representative models (weak/mid/strong). Inner ring: pass/syntax/semantic split (percentages shown). Outer ring: syntax errors split into type/sort mismatches and API/reference errors; semantic errors split into domain pattern errors and translation logic errors. \ours\ eliminates type/sort errors and reduces API errors across all tiers. Opus baseline failures are almost exclusively domain pattern errors; \ours\ resolves most of them.}
    \label{fig:errors}
\end{figure*}

Fig.~\ref{fig:errors} breaks down the 245 tasks into three categories for each model configuration: pass, syntax error (specification does not execute due to Z3 type mismatches, undefined functions, or Python errors) and semantic error (specification executes but verification fails due to incorrect logic). 
We select three representative models spanning weak, mid-tier, and strong baselines. 
The inner ring shows the three-way split (pass, syntax, semantic); 
the outer ring further divides syntax errors into \textit{type/sort mismatches} (Z3 bit-vector size or type conflicts) and \textit{API/reference errors} (undefined Z3 functions or attributes), and semantic errors into \textit{domain pattern errors} (failures in syscalls requiring specialized patterns such as IPC, PTE formulas, or IOMMU mappings) and \textit{translation logic errors} (incorrect condition negation or field mapping in structurally simpler syscalls).

\textbf{Strong models} (Claude Opus 4.6, Devstral 2, Claude Sonnet 4, Claude Haiku 4.5): 
\ours\ substantially reduces both error types, as shown in panels (b) and (c). 
Devstral 2's syntax errors drop from 31.0\% to 8.2\%; the outer ring decomposes this reduction into two components: type/sort mismatches are completely eliminated (6.1\% $\to$ 0\%), while API/reference errors decrease from 24.9\% to 8.2\%. 
For Claude Opus 4.6, the baseline semantic errors are almost exclusively domain pattern errors, with translation logic errors nearly absent (0.4\%), indicating that Claude Opus 4.6 already handles simple translations well. 
\ours\ reduces domain pattern errors to 2.4\%, leaving only 8 failures.

\textbf{Weaker models} (Nova Premier, DeepSeek-V3.2, Llama 4 Maverick, Qwen3): 
\ours\ reduces both error types, but substantial residual errors remain, as illustrated by Qwen3 in panels (a). 
Unlike strong models, Qwen3's semantic errors are split nearly equally between domain pattern and translation logic, indicating that weaker models struggle with both specialized patterns and basic translation. 
\ours\ provides moderate reductions in both sub-categories, but 30.2\% semantic errors persist.

\textbf{DeepSeek-R1} represents a special case: 
as a reasoning-optimized model, it exhibits the highest syntax error rate (59.2\% baseline, 53.1\% with \ours). 
Manual inspection reveals that DeepSeek-R1 frequently fails to produce output in the required \texttt{```python```} format, instead generating reasoning in natural language. 
Its low semantic error rate suggests that when DeepSeek-R1 does produce valid specifications, they tend to be semantically reasonable.

\textbf{Residual failure analysis:} 
We examine the 8 remaining Claude Opus 4.6 + \ours\ failures. 
The failures concentrate on two system calls:
\texttt{sys\_reclaim\_intremap} fails in all 5 variants (\textit{semantic errors}): 
this system call handles the reclamation of the interrupt mapping table, coordinating interactions among the IOMMU state, the interrupt vectors and the ownership of the device. 
Neither baseline nor \ours\ produces a correct specification, suggesting that the required pattern (coordinating multiple hardware subsystem states during resource reclamation) exceeds the guide's coverage.
\texttt{sys\_lseek} fails in 1 variant (\textit{syntax error}): 
the model generates \texttt{z3.SLT} for a signed comparison, but Z3 v4.5.0 does not expose this function. 
The remaining 2 failures occur in \texttt{sys\_alloc\_intremap} and \texttt{sys\_alloc\_io\_bitmap}.
(\textit{semantic errors}) The concentration of most failures in the interrupt remapping subsystem suggests a clear direction for guide improvement.

\textbf{Z3 verifier feedback:}
When a specification fails, the Z3 verification pipeline produces diagnostically distinct feedback depending on the error class.
For \textit{syntax errors}, the Python/Z3 runtime raises an exception (e.g. \texttt{TypeError}, \texttt{Z3Exception}) with a message that pinpoints the fault. Messages such as \texttt{Sort mismatch: expected BitVec(64), got BitVec(32)} directly identify the offending field and the expected type.
This feedback is precise and machine-readable, making it suitable for automated repair: an LLM could receive the traceback and re-generate only the erroneous expression.
For \textit{semantic errors}, the specification executes successfully but Z3 finds a concrete $\mathbf{a}^*$ violating Eq.~\ref{eq:verify} and produces a \textit{counterexample}: an assignment of bitvector values to all symbolic inputs, such as \texttt{pid = 0$\times$3} and \texttt{ppid = 0$\times$1}, under which the implementation and specification diverge.
While counterexamples formally prove incorrectness, interpreting them requires mapping raw bitvector values back to domain-level semantics (process states, page table entries), a non-trivial task that limits their direct utility for automated repair.
This asymmetry, actionable feedback for syntax errors but opaque feedback for semantic errors, suggests that a verification-feedback repair loop (Section~\ref{sec:conclusion}) would be most effective for syntax failures, where the error signal is both precise and self-contained.

\begin{table*}[htbp]
    \caption{Selected system calls showing \ours\ impact (Claude Opus 4.6, Pass@1 across 5 variants per syscall).}
    \begin{center}
        \begin{tabular}{l ccc l}
            \toprule
            \textbf{System Call} & \textbf{Category} & \textbf{Baseline} & \textbf{+ \ours} & \textbf{Guide sections used}\\
            \midrule
            \texttt{call\_proc} & IPC & 0/5 & 5/5 & IPC patterns\\
            \texttt{send\_proc} & IPC & 0/5 & 5/5 & IPC patterns\\
            \texttt{reply\_wait\_proc} & IPC & 0/5 & 5/5 & IPC patterns\\
            \texttt{sys\_alloc\_iommu\_pt} & IOMMU alloc & 0/5 & 5/5 & PTE formulas, shadow metadata\\
            \texttt{sys\_alloc\_iommu\_pdpt} & IOMMU alloc & 1/5 & 5/5 & PTE formulas, field mapping\\
            \texttt{sys\_reclaim\_page} & Page reclaim & 2/5 & 5/5 & Refcnt, state pointers\\
            \texttt{sys\_reclaim\_iommu\_frame} & IOMMU reclaim & 2/5 & 5/5 & PTE formulas, refcnt\\
            \texttt{sys\_alloc\_intremap} & Interrupt & 0/5 & 4/5 & Field mapping, helpers\\
            \texttt{sys\_lseek} & File management & 5/5 & 4/5 & (regression)\\
            \texttt{sys\_map\_pml4} & Page mapping & 5/5 & 5/5 & (already correct)\\
            \bottomrule
        \end{tabular}
        \label{tab3}
    \end{center}
\end{table*}

\subsection{Model vs. Method Contribution}

With nine models spanning six providers, we can now robustly disentangle the model capability from the \ours\ contribution.
The mean improvement with \ours\ is $+$18.7\%, with all nine models showing statistically significant gains in a controlled comparison (same model, same tasks, identical evaluation protocol).

Three observations emerge:

\begin{itemize}

    \item The contribution of \ours\ is \textit{model-agnostic}:
    it improves models regardless of provider (Anthropic, Mistral, Amazon, DeepSeek, Meta, Alibaba), architecture (dense, MoE) or capability level.
    This rules out the hypothesis that the effectiveness of guide is specific to a particular model family's training data.

    \item The effect of \ours\ is \textit{comparable to or greater than a full model generation upgrade}: 
    For example, the improvement from the Claude Sonnet 3.5 baseline to the Claude Sonnet 4 baseline is $+$16.3\%, while applying \ours\ to Sonnet 4 yields an additional $+$13.9\%.
    For Devstral 2, \ours\ alone ($+$31.8\%) exceeds the effect of any single model upgrade observed.

    \item The relationship between baseline capability and \ours\ gain is \textit{non-linear}:
    The greatest gains are accrued to mid-tier models (Mistral Devstral 2: $+$31.8\%, Claude Haiku 4.5: $+$26.1\%) rather than the strongest or weakest.
    We hypothesize that \ours\ requires sufficient instruction-following ability to parse and apply the guide's patterns, but yields the largest marginal gains when the model's intrinsic domain knowledge is incomplete, a condition more characteristic of mid-tier models.
    
\end{itemize}

\subsection{Per-Syscall Analysis}

To understand where the guide helps the most, we analyze the 49 system calls (each appearing in 5 task variants) in OSV-Bench \cite{b1}. 
We take Claude Opus 4.6 as an example: TABLE~\ref{tab3} shows representative system calls with notable improvement patterns.

The greatest improvements occur in system calls that require domain-specific translation patterns. 
All three IPC system calls improve from 0/5 to 5/5, as the guide provides dedicated IPC state-machine patterns that are difficult to infer from the few-shot examples alone. 
Similarly, IOMMU-related system calls benefit substantially from the PTE formula and shadow metadata categories. 
System calls with straightforward logic show no improvement, as the baseline already succeeds. 
A minor regression occurs for \texttt{sys\_lseek}, caused by a mismatch in the Z3 API version, as discussed in the residual failure analysis.

\subsection{Qualitative analysis}

To illustrate how \ours\ changes model behavior concretely, we examine three representative cases where the baseline (Claude Opus 4.6) fails but \ours\ succeeds.

\textbf{Case 1: IPC state machine logic:} 
The \texttt{call\_proc} system call requires checking whether the receiver process is willing to accept messages from the current sender: 

\begin{tcolorbox}[enhanced, boxrule=0.4pt, arc=2pt, left=4pt, right=4pt, top=2pt, bottom=2pt,
    colframe=red!60!black, colback=red!3, colbacktitle=red!60!black,
    coltitle=white, fonttitle=\sffamily\bfseries\footnotesize,
    toptitle=1.5pt, bottomtitle=1.5pt, title={\ding{55}\ Baseline: disjunctive check}]
    \ttfamily\footnotesize
    \textcolor{codeyellow}{z3.Or}(\\
    \quad old.procs[pid].ipc\_from \textcolor{codeblue}{==} \textcolor{codeyellow}{z3.BitVecVal}(0, \textcolor{codecyan}{dt}.pid\_t),\\
    \quad old.procs[pid].ipc\_from \textcolor{codeblue}{==} old.current),\quad \textcolor{codegreen}{\# wrong semantics}
\end{tcolorbox}\vspace{-2pt}

\begin{tcolorbox}[enhanced, boxrule=0.4pt, arc=2pt, left=4pt, right=4pt, top=2pt, bottom=2pt,
    colframe=green!50!black, colback=green!3, colbacktitle=green!50!black,
    coltitle=white, fonttitle=\sffamily\bfseries\footnotesize,
    toptitle=1.5pt, bottomtitle=1.5pt, title={\ding{51}\ \ours: implication check}]
    \ttfamily\footnotesize
    \textcolor{codeyellow}{z3.Implies}(old.procs[pid].ipc\_from \textcolor{codered}{!=} 0,\\
    \quad\quad\quad\quad\quad\, old.procs[pid].ipc\_from \textcolor{codeblue}{==} old.current),
\end{tcolorbox}

\noindent The baseline conflates two independent conditions into a disjunction, which allows for invalid states. 
The IPC pattern section of \ours\ specifies the correct implication semantics: if the receiver has a designated sender (\texttt{ipc\_from != 0}), it must match the current process.

\textbf{Case 2: IOMMU page table entry formula:} 
For \texttt{sys\_alloc\_iommu\_pt}, the baseline applies the x86 PTE formula to an IOMMU entry:

\begin{tcolorbox}[enhanced, boxrule=0.4pt, arc=2pt, left=4pt, right=4pt, top=2pt, bottom=2pt,
    colframe=red!60!black, colback=red!3, colbacktitle=red!60!black,
    coltitle=white, fonttitle=\sffamily\bfseries\footnotesize,
    toptitle=1.5pt, bottomtitle=1.5pt, title={\ding{55}\ Baseline: x86-style PFN shift}]
    \ttfamily\footnotesize
    pfn = \textcolor{codeyellow}{z3.UDiv}(old.pages\_ptr\_to\_int,\\
    \quad\quad\quad\quad\quad\quad\, \textcolor{codeyellow}{z3.BitVecVal}(\textcolor{codecyan}{dt}.PAGE\_SIZE, 64)) + to\\
    \textcolor{codeblue}{new}.pages[frm].data[index] = (pfn \textcolor{codered}{<<} \textcolor{codecyan}{dt}.DMAR\_PTE\_ADDR\_SHIFT) | perm\quad \textcolor{codegreen}{\# wrong formula}
\end{tcolorbox}\vspace{-2pt}

\begin{tcolorbox}[enhanced, boxrule=0.4pt, arc=2pt, left=4pt, right=4pt, top=2pt, bottom=2pt,
    colframe=green!50!black, colback=green!3, colbacktitle=green!50!black,
    coltitle=white, fonttitle=\sffamily\bfseries\footnotesize,
    toptitle=1.5pt, bottomtitle=1.5pt, title={\ding{51}\ \ours: IOMMU byte-addressed}]
    \ttfamily\footnotesize
    \textcolor{codeblue}{new}.pages[frm].data[index] =\\
    \quad (\textcolor{codeblue}{new}.pages\_ptr\_to\_int + to * \textcolor{codecyan}{dt}.PAGE\_SIZE) | perm
\end{tcolorbox}

\noindent The x86 page tables encode entries as shifted page-frame numbers, but IOMMU entries use direct byte addresses. 
Without the explicit distinction of the guide between these two formula families (Category 7 in Table~\ref{tab1}), the model defaults to the more common x86 pattern.

\textbf{Case 3: Map field read/write syntax:} 
A common syntax error involves Z3's dual-access convention for map fields:

\begin{tcolorbox}[enhanced, boxrule=0.4pt, arc=2pt, left=4pt, right=4pt, top=2pt, bottom=2pt,
    colframe=red!60!black, colback=red!3, colbacktitle=red!60!black,
    coltitle=white, fonttitle=\sffamily\bfseries\footnotesize,
    toptitle=1.5pt, bottomtitle=1.5pt, title={\ding{55}\ Baseline: parentheses for write}]
    \ttfamily\footnotesize
    \textcolor{codeblue}{new}.pages[pn].data\textcolor{codered}{\textbf{(}}index\textcolor{codered}{\textbf{)}} = value\quad \textcolor{codegreen}{\# TypeError}
\end{tcolorbox}\vspace{-2pt}

\begin{tcolorbox}[enhanced, boxrule=0.4pt, arc=2pt, left=4pt, right=4pt, top=2pt, bottom=2pt,
    colframe=green!50!black, colback=green!3, colbacktitle=green!50!black,
    coltitle=white, fonttitle=\sffamily\bfseries\footnotesize,
    toptitle=1.5pt, bottomtitle=1.5pt, title={\ding{51}\ \ours: brackets for write}]
    \ttfamily\footnotesize
    \textcolor{codeblue}{new}.pages[pn].data\textcolor{codeblue}{\textbf{[}}index\textcolor{codeblue}{\textbf{]}} = value
\end{tcolorbox}

\noindent In the Z3 Python API, the map fields use parentheses \texttt{()} for reads, but brackets \texttt{[]} for writes. 
This dual convention is not documented in the programming model document provided in the prompt (Section~\ref{sec:prompt}) in OSV-Bench \cite{b1}, making it a frequent source of syntax errors.
The Category 4 of \ours\ (Table~\ref{tab1}) explicitly lists all dual-access fields with both access patterns.

These cases illustrate the three error classes addressed by \ours: semantic logic errors (Case 1), domain knowledge gaps (Case 2) and syntax convention errors (Case 3).

\section{Related Work}

\textbf{LLM-based generation of formal artifacts:}
Code2Inv \cite{b12} pioneered neural approaches to formal reasoning with reinforcement learning for loop invariant inference.
Baldur \cite{b13} applied LLMs to whole-proof generation in Isabelle/HOL.
SpecGen \cite{b14} explored LLM-based specification generation for general-purpose programs.
More recently, LLMs have been applied to generate other formal artifacts: 
IRIS \cite{b18} combines LLMs with static analysis to synthesize security vulnerability detectors, 
and QLCoder \cite{b19} uses LLMs to generate CodeQL queries for static analysis.
These works share a common theme with \ours: LLMs generating domain-specific formal code (Z3 specifications, CodeQL queries, Isabelle proofs) where general code generation capability is insufficient and structured domain knowledge is required.

\textbf{Neurosymbolic programming:}
Neurosymbolic method \cite{swarat2021neurosymbolic, zhu2025locus} concerns combining neural modules along with symbolic algorithms on downstream applications. 
Existing frameworks like \cite{b21, biberstein2025lobster, li2024relational} introduce programming languages that integrate neural computation with symbolic reasoning, enabling neural networks to leverage logical rules during inference.
\ours\ operates at a different level of abstraction: rather than embedding symbolic reasoning into the model architecture, we inject domain knowledge into the prompt, allowing unmodified LLMs to produce formally verifiable artifacts.
Both approaches address the fundamental challenge of bridging neural and symbolic reasoning, but \ours\ requires no model modification or training.

\textbf{Prompting methods for code generation:}
Chain-of-Thought prompting \cite{b10} and its structured variant SCoT \cite{b2} improve code generation by generating intermediate reasoning.
\ours\ adapts SCoT principles to specification generation by providing structured domain knowledge rather than eliciting reasoning chains: the structure is in the reference material, not the output.

\textbf{OS kernel verification:}
Hyperkernel \cite{b7} introduced push-button verification for an OS kernel using Z3.
SeL4 \cite{b4} demonstrated comprehensive formal verification of a microkernel.
OSV-Bench \cite{b1} established the first LLM benchmark for this domain.
Our work builds on OSV-Bench by demonstrating that domain-specific prompting can substantially improve LLM performance on specification generation.

\textbf{Knowledge-enhanced prompting:}
Retrieval-augmented generation (RAG) \cite{b15} and tool-augmented approaches inject external knowledge into LLM prompts.
Chen et al.\ \cite{b20} study how API documentation aids LLM code generation, finding that example code in documentation contributes more than descriptive text --- a finding consistent with our guide design, which emphasizes concrete translation patterns over abstract rules.
Our translation guide serves a similar purpose to RAG but is hand-crafted for the specific domain, providing higher precision than retrieved passages.
The guide can be viewed as a form of structured few-shot documentation: more comprehensive than examples alone, but more targeted than a full manual.

\section{Discussion}

\subsection{Threats to Validity}

\textbf{Internal validity:}
The translation guide was constructed using failure analysis on Claude Haiku 4.5, which is also one of the nine evaluated models.
This raises the concern that improvements on Claude Haiku 4.5 may partly reflect overfitting to the development model's failure modes rather than general translation knowledge.
We mitigate this in two ways: 
\textit{(1)} the guide contains only general translation rules and illustrative patterns, not task-specific solutions; 
and \textit{(2)} the largest improvement is observed on Devstral 2 ($+$31.8\%), a model from a different provider that was not involved in guide development, demonstrating cross-model generalization.
A second concern is potential oracle knowledge leakage: the guide patterns were derived from inspecting oracle specifications.
However, the guide encodes only reusable domain rules (e.g. IOMMU PTEs use byte addressing, not 4K-page addressing) rather than complete specifications for any task.
The 15 categories cover general patterns applicable across multiple system calls, functioning as a domain textbook rather than an answer key.

\textbf{External validity:}
Our evaluation is conducted on a single benchmark (OSV-Bench) targeting a single codebase (Hyperkernel)\cite{b1}.
The 245 tasks span 49 system calls within one x86 microkernel, and the translation guide is specific to the Hyperkernel programming model and Z3 specification language.
Whether the guide-based prompting approach generalizes to other verification targets, specification languages, or OS kernels remains an open question.
However, the methodology, systematic failure analysis followed by structured knowledge injection, is not inherently domain-specific and could be applied to other formal verification tasks.

\textbf{Construct validity:}
We use Pass@1 as the primary metric, which is binary: a specification either passes or fails Z3 verification.
This metric does not capture partial correctness. For example, a specification that correctly translates 95\% of a system call's logic but mishandles one edge case receives the same score as one with entirely incorrect logic.
The three-way error classification (pass/syntax/semantic) partially addresses this limitation by distinguishing failure modes.
Additionally, we report single-run results with greedy decoding rather than averaging multiple runs, which eliminates sampling variance for most models but means our results reflect a single deterministic configuration rather than expected performance across temperature settings.

\subsection{Limitations}

\textbf{Guide construction cost:}
The translation guide required analyzing 49 oracle specifications to identify patterns.
This one-time effort does not generalize to new verification domains without corresponding analysis.

\textbf{Output format dependency:}
\ours\ assumes that the model can produce syntactically valid Python in the requested format.
Models that struggle with format compliance (e.g. DeepSeek-R1, which outputs reasoning traces rather than code blocks in 53-59\% of tasks) derive limited benefit.
A preprocessing or re-prompting step to enforce output format could improve results for such models.

\textbf{Single-call vs. multi-stage:}
\ours\ injects the guide into a single LLM call.
A multi-stage approach that first analyzes the system call and then generates pre-conditions and post-conditions separately might yield further improvements, particularly for models with smaller context windows.

\section{Conclusion}\label{sec:conclusion}

We demonstrate that \ours, a domain knowledge prompting method with structured translation guides, significantly improves LLM performance in formal specification generation across a broad range of models.
Evaluated on nine models from six providers, \ours\ improves every model tested, with a mean gain of $+$18.7 percentage points and the best configuration reaching 96.73\% Pass@1, substantially exceeding the previous best of 55.10\%.
The key insight is that specification generation requires domain-specific translation knowledge (page table formulas, reference counting semantics, field naming conventions) that cannot be reliably inferred from few-shot examples alone but can be effectively communicated through a structured reference document.

The model-agnostic effectiveness of \ours\ across diverse architectures (dense, mixture-of-experts, reasoning) and providers demonstrates that the technique is not tied to any particular model family.
The largest gains accrue to mid-tier models with sufficient instruction-following capability but incomplete domain knowledge, pointing to a productive division of labor between human domain experts (who construct the guide) and LLMs (who apply it at scale).

Future work includes:
\textit{(1)} exploring multi-stage generation where pre-conditions and post-conditions are generated in separate LLM calls,
\textit{(2)} adding a verification-feedback repair stage to address remaining failures,
and \textit{(3)} extending to other verification domains to test transferability of the \ours\ framework.
\bibliographystyle{IEEEtran}
\bibliography{references}

\end{document}